\pdfoutput=1

\documentclass[10pt,twocolumn,letterpaper]{article}

\usepackage[pagenumbers]{cvpr}

\usepackage{algorithm}
\usepackage{algpseudocode}

\usepackage[most]{tcolorbox}

\newtcolorbox{promptbox}[1]{
  enhanced,
  colback=viderlight!35,
  colframe=viderpurple!85,
  coltitle=white,
  fonttitle=\bfseries\small\sffamily,
  title=#1,
  boxrule=0.6pt,
  arc=2.5pt,
  left=7pt, right=7pt, top=6pt, bottom=6pt,
  toptitle=3pt, bottomtitle=3pt,
  before skip=6pt, after skip=6pt,
  fontupper=\small\sffamily\raggedright,
}
\newcommand{\phtag}[1]{\texttt{\textcolor{viderpurple}{#1}}}

\usepackage{colortbl}
\usepackage{multirow,tabularx}

\definecolor{viderpurple}{HTML}{75649B}
\definecolor{viderlight}{HTML}{F1EEF6}

\definecolor{fighlred}{HTML}{E8A0AA}
\definecolor{fighlyellow}{HTML}{FFE6A3}
\definecolor{fighlgreen}{HTML}{7FC9A3}
\DeclareRobustCommand{\fighl}[2]{{\setlength{\fboxsep}{1pt}\colorbox{#1}{#2}}}

\definecolor{cvprblue}{rgb}{0.21,0.49,0.74}
\usepackage[pagebackref,breaklinks,colorlinks,allcolors=cvprblue]{hyperref}

\def\confName{CVPR}
\def\confYear{2027}

\title{Same Answer, Different Evidence: Visual Dependence Rewards for Test-Time Reinforcement Learning}

\author{%
    Zhen Zhang\textsuperscript{1,}\thanks{Equal contribution.}\quad
    Wenfang Sun\textsuperscript{2,}\footnotemark[1]\quad
    Cheng Li\textsuperscript{3}\quad
    Le Wu\textsuperscript{1,}\thanks{Corresponding author.}\\
    Hefei Xu\textsuperscript{1}\quad
    Yingjun Du\textsuperscript{2}\quad
    Min Hou\textsuperscript{1}\quad
    Cees G. M. Snoek\textsuperscript{2}\\[4pt]
    {\normalsize
        \textsuperscript{1}Hefei University of Technology\quad
        \textsuperscript{2}University of Amsterdam\quad
        \textsuperscript{3}Jianghuai Advance Technology Center}\\[6pt]
    {\normalsize\href{https://zhangzhen258.github.io/ViDeR/}{\textbf{Project Page \& Code}}}
}

\begin{document}

\maketitle

\begin{abstract}
Consensus-based test-time reinforcement learning adapts vision-language models on unlabeled test data by rewarding agreement among sampled answers. Yet responses with the same final answer can differ in their reliance on visual evidence: some draw on the image, while others arrive at that answer primarily through language priors. Answer-consensus rewards do not distinguish between them.
Our key insight is that visual dependence should be tested on the evidence each response actually uses rather than on the image as a whole: attention indicates where a response's reasoning looks, and occluding only those regions while keeping the rest of the image reveals whether that reasoning needs them.
We propose Visual-Dependence Reward (ViDeR), which augments answer consensus with a response-specific measure of sensitivity to visual evidence. Rather than perturbing the entire image, ViDeR uses each response's reasoning-token attention to identify candidate evidence and selectively occludes the corresponding regions. It then scores the same response under the original and occluded images, measuring the attention-weighted decrease in token likelihood.
This score estimates how strongly the response depends on the selected visual evidence and modulates its answer-frequency reward. Consequently, responses with identical final answers can receive different learning signals, without requiring answer labels or region annotations. Across eight benchmarks and six backbones, ViDeR achieves the highest average benchmark score on each backbone among the evaluated methods.
\end{abstract}

\section{Introduction}
\label{sec:intro}

\begin{figure}[!t]
    \centering
    \includegraphics[width=\linewidth]{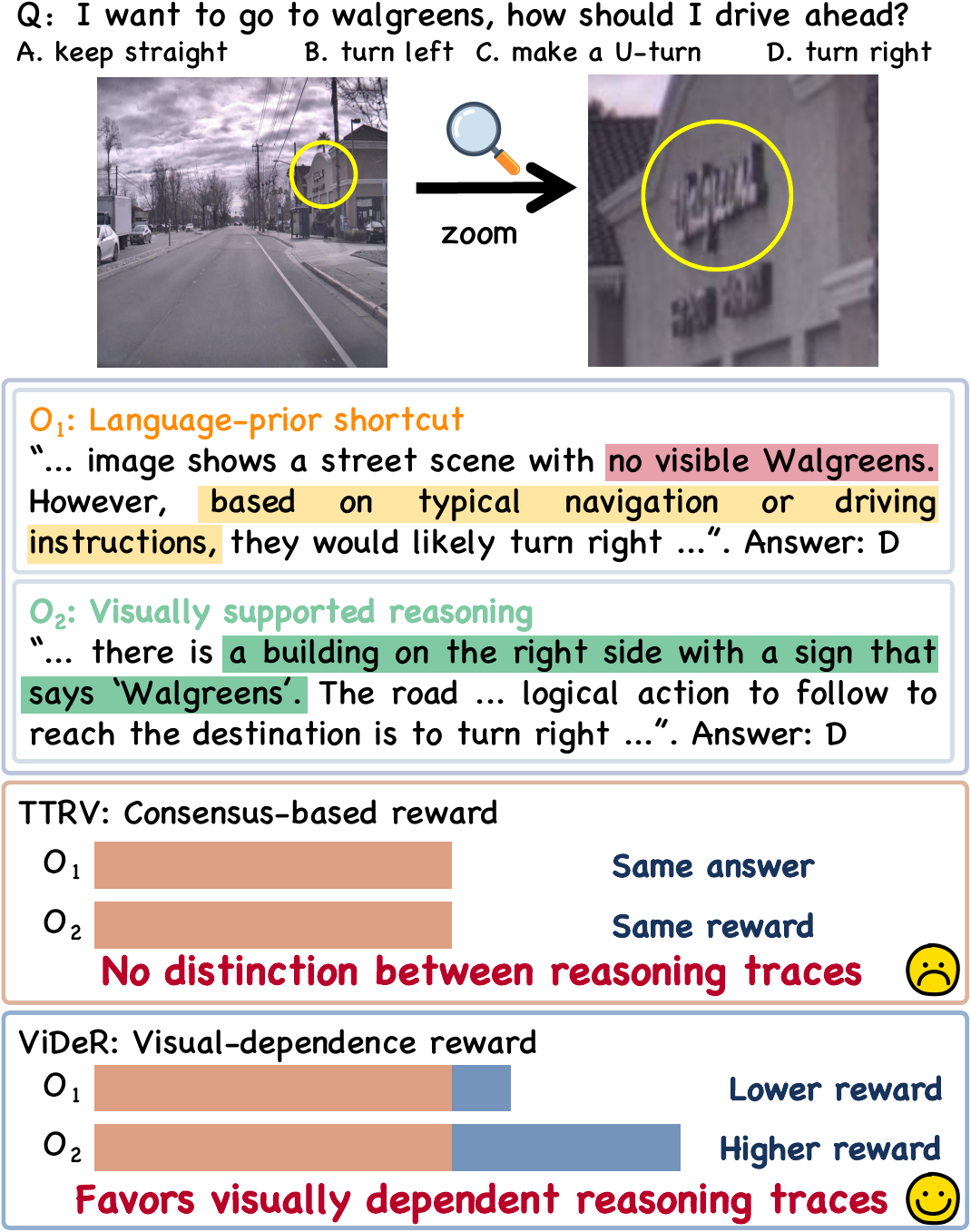}
    \caption{\textbf{Same answer, different evidence.} Both responses select same answer, but $O_1$ denies that Walgreens is visible and relies on typical navigation (language priors), whereas $O_2$ cites the store sign on the right (visual dependence).
TTRV~\cite{TTRV} assigns both the same reward since they have the same final answer.
ViDeR adds a  visual-dependence bonus (blue) to the answer-frequency term (peach), favoring reasoning that depends more on visual evidence.
Highlights mark misperception (\fighl{fighlred}{red}), prior-driven reasoning (\fighl{fighlyellow}{yellow}), and visually supported reasoning (\fighl{fighlgreen}{green}).}
    \label{fig:motivation}
\end{figure}

Vision--language models (VLMs) have become capable visual reasoners~\cite{mPLUGOwl22024,Monkey2024,QACap2025,InternVL2024,Cambrian12024}, yet they remain static once trained. When deployed on a new domain or task, improving them further usually requires annotated data and additional training~\cite{SFT,GRPO}.
Test-time reinforcement learning methods avoid this cost by adapting the model on unlabeled test inputs.
Unlike test-time scaling, which spends extra inference compute without updating the model~\cite{Snell2024,SelfConsistency,LetsVerify}, they turn the model's own samples into a training signal: TTRL~\cite{TTRL} rewards responses that agree with the majority-voted answer, follow-up work refines this consensus signal~\cite{SCOPE,SelfHarmony,SR-TTRL}, and TTRV~\cite{TTRV} extends it to VLMs with a reward based on answer frequency.

These methods share a common limitation: the reward depends only on the final answer, regardless of the reasoning that produces it.
Yet responses with the same final answer can differ in how much they rely on visual evidence.
As illustrated in \Cref{fig:motivation}, $O_1$ and $O_2$ reach the same answer through different reasoning paths.
$O_1$ misperceives the image and selects a plausible option based on language priors, whereas $O_2$ identifies the relevant visual evidence and derives the answer from it.
Because their final answers agree, answer-consensus rewards assign both responses the same reward and cannot separate visually grounded reasoning from prior-driven reasoning.
During adaptation, prior-driven responses are thus reinforced as strongly as grounded ones whenever they reach the consensus answer.
A natural remedy is to measure how much each response depends on the image.
TTIQ~\cite{TTIQ} takes a step in this direction by re-scoring responses with the image replaced by an all-zero tensor, but removing the whole image discards all visual context at once and cannot identify which evidence a particular response relies on.

In this paper, we start from a counterfactual premise: if a response derives its conclusion from the image, removing the visual evidence it relies on should make that same response less likely, whereas a response driven by language priors should be largely unaffected. Our key insight is that visual dependence should be tested on the evidence each response actually uses: attention indicates where a response's reasoning looks, and occluding only those regions while keeping the rest of the image reveals whether the reasoning needs them. Unlike whole-image removal, this test is specific to each response and preserves the remaining visual context, so it isolates the evidence a particular trace attends to rather than testing whether the response depends on having an image at all. Answer consensus asks whether responses converge on the same answer, whereas this test asks whether a reasoning trace needs the evidence it looks at. Training-time methods exploit related signals through image masking or explicit perception rewards~\cite{PAPO,VisionaryR1,PEARL}, but they build on rewards from ground-truth answers. As our test is read from the model's own likelihood, it requires no labels and serves as a reward on unlabeled test data.

We propose Visual-Dependence Reward (ViDeR), which augments answer consensus with a response-specific measure of sensitivity to visual evidence. For each sampled response, ViDeR aggregates the attention that its reasoning tokens direct toward the image to identify candidate evidence, and selectively occludes the corresponding regions. Attention alone, however, reveals only where the model looked, not whether its reasoning needed what it saw. ViDeR therefore scores the same response under the original and the occluded image without regenerating it, and aggregates the token-level likelihood decreases, weighted by each token's attention to the occluded regions, into a score that estimates how strongly the response depends on the selected evidence. This score modulates the answer-consensus reward by adding a bonus proportional to it, so responses with the same answer but different evidence can receive different learning signals: frequency carries consensus at the answer level, while visual dependence evaluates each reasoning trace on its own. Neither answer labels nor region annotations are required. Our contributions are:
\begin{itemize}
    \item We identify a limitation of answer-consensus rewards for VLMs: they reward the same answer equally, whether based on visual evidence or language priors.
    \item We propose ViDeR, a label-free reward that measures each response's visual dependence through attention-guided occlusion and uses this signal to modulate the answer-consensus reward.
    \item Across eight multimodal benchmarks and six VLM backbones, ViDeR achieves the highest average score on every backbone among the evaluated methods, and comparisons with whole-image and random interventions indicate that the choice of occluded regions matters.
\end{itemize}

\begin{figure*}[!t]
    \centering
    \includegraphics[width=\textwidth]{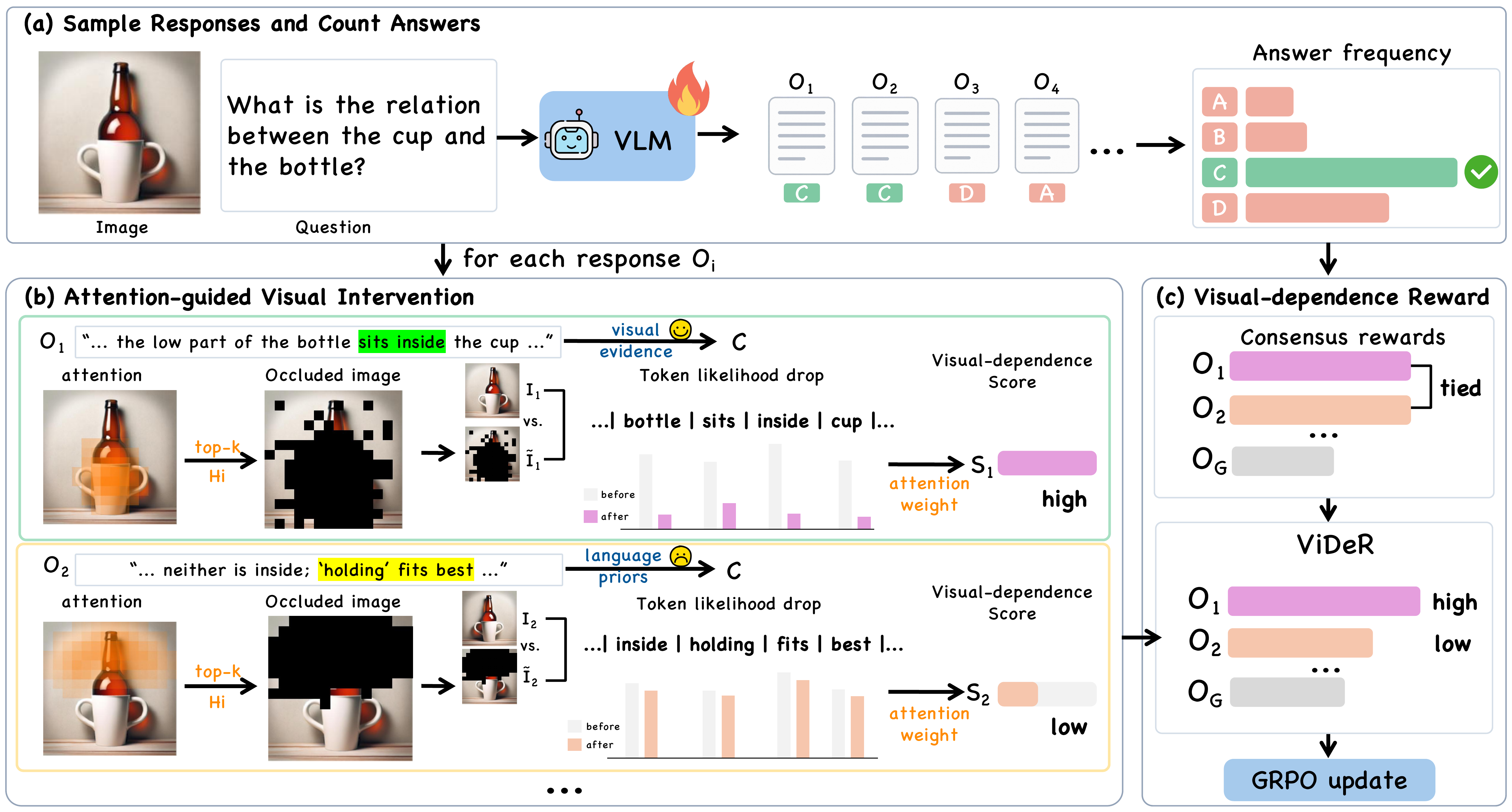}
    \caption{\textbf{Overview of ViDeR.} (a) The VLM samples multiple responses, from which answer frequencies are computed. (b) Attention-guided visual intervention masks highly attended image regions for each response. The same reasoning tokens are evaluated on the original and masked images while keeping their textual prefixes fixed. Confidence-calibrated token scores derived from the likelihood drops are aggregated with attention weights to obtain a visual-dependence score. (c) ViDeR combines this score with answer frequency to construct rewards for GRPO updates. $O_1$ and $O_2$ produce the \textbf{same answer} C but rely on \textbf{different evidence}.  $O_1$ exhibits stronger visual dependence and therefore receives a higher ViDeR reward than $O_2$.}
    \label{fig:framework}
\end{figure*}

\section{Related Work}

\noindent\textbf{Multimodal Reinforcement Learning.}
Verifiable task rewards support RL fine-tuning for visual perception and
multimodal reasoning~\cite{VisualRFT,MMEureka,VLMR1,VisionR1,R1Onevision}.
Beyond reward design, other methods improve how responses are sampled during RL,
by prompting the model to reflect on its answers or by perturbing the input
image to diversify the rollouts~\cite{VLRethinker,NoisyRollout}.
Perception-oriented methods elicit image descriptions, reward perceptual
accuracy, or assess whether generated descriptions suffice for
reasoning~\cite{VisionaryR1,PerceptionR1,RegionReasoner,PEARL,VisionSR1},
though the benefits of perception and reasoning rewards vary across
tasks~\cite{PerceptionReasoningReward}.
Process supervision guides attention with annotated evidence or verifies visual
claims with learned reward models~\cite{RememberR1,EASE,Perceval}, and evidence
rewards use an LLM judge to check that reasoning refers to visual evidence
extracted by a larger model~\cite{VER}.
Another line of work makes visual evidence acquisition explicit, grounding
intermediate steps in image coordinates or gathering targeted observations
by zooming into image regions~\cite{ViGoRL,DeepEyes,PixelReasoner}.
DeFacto~\cite{DeFacto} combines counterfactual image masking
with rewards for answer correctness and evidence-region selection.
Tool-using policies learn to call external vision tools or execute
image-processing code~\cite{OpenThinkIMG,VToolR1,Thyme}, and sufficiency and
compactness rewards encourage the use of relevant visual evidence gathered
during reasoning~\cite{FaithfulVisualThinking}.
These methods show that visual signals complement answer-level rewards. However, they rely on training-time objectives that require supervision beyond the model itself, such as ground-truth answers or annotated regions. We address whether visual reliance can serve as a reward signal on unlabeled test data, where ground-truth answers are unavailable.
ViDeR obtains this signal from the model itself: in the spirit of
occlusion-based sensitivity analysis~\cite{ZeilerFergus}, it occludes the image
regions that a sampled reasoning trace attends to and uses the resulting
likelihood drop of that trace as a visual reward, requiring neither
ground-truth answers nor visual annotations.

\noindent\textbf{Test-Time Reinforcement Learning.}
Test-time reinforcement learning adapts models on unlabeled inputs using
rewards constructed at test time. TTRL~\cite{TTRL} turns answer consensus into
a training signal through majority-vote pseudo-labels. Subsequent work refines
this signal with entropy and confidence~\cite{ETTRL,ECHO,RENT,SCRL,SCOPE,COMPASS},
agreement across question reformulations~\cite{SelfHarmony},
self-reflection~\cite{SR-TTRL,TTSR}, and distribution- or graph-aware
consensus~\cite{DARE,CoRE}, or extends it beyond answer matching with
execution feedback and self-generated rubrics~\cite{T3RL,ERPO,SERPO}.
These methods establish consensus among the model's own samples as an
effective label-free training signal.
For vision--language tasks, answer-consensus rewards capture agreement among
sampled answers on images or video frame subsets~\cite{TTRV,TTAVid,TTRLACU}.
Spatial agreement rewards GUI grounding~\cite{GUIRCPO}, and alignment and
dispersion rewards guide visual encoder adaptation for
recognition~\cite{GRPOTTA}.
These rewards are computed from final outputs or textual confidence and do not
explicitly consider how the reasoning uses the image; under answer-consensus
rewards, for example, responses that reach the same answer receive the same
reward even if some rely on prior knowledge rather than on the image, a
shortcut long documented as language priors in visual question
answering~\cite{VQAv2,VQACP}.
TTIQ~\cite{TTIQ} takes a step toward measuring visual dependence by re-scoring
responses with the whole image or the question removed.
Yet this reflect reliance on the image as a
whole rather than on the specific regions a response draws on.
ViDeR measures this dependence at the region level: it occludes the regions
that each response's reasoning attends to while keeping the remaining visual
context, and uses the resulting likelihood drop to modulate the
answer-frequency reward.

\section{Preliminaries}

\paragraph{Test-Time Reinforcement Learning.}
Given an unlabeled target dataset $\mathcal{D}_{\mathrm{test}}=\{(I_n,q_n)\}_{n=1}^{N}$ of image--question pairs, test-time reinforcement learning adapts a pretrained VLM $\pi_{\theta_0}$ using rewards computed from its own responses. For each pair $(I,q)$, the rollout policy $\pi_{\theta_{\mathrm{old}}}$ samples a group of $G$ responses $\mathcal{O}=\{o_i\}_{i=1}^{G}$. Each response $o_i$ consists of a reasoning trace, whose token positions are denoted $\mathcal{T}_i$, followed by a final answer $a_i$. We update the policy with GRPO~\cite{GRPO}, which normalizes rewards within each group to obtain advantages.

\paragraph{Answer-Consensus Reward.}
Answer-consensus methods~\cite{TTRL,TTRV} treat agreement among sampled answers as a signal of correctness. Following TTRV~\cite{TTRV}, we use the frequency of an answer within the group as its reward:
\begin{equation}
    p(a_i)=\frac{1}{G}\sum_{j=1}^{G}\mathbf{1}[a_j=a_i].
    \label{eq:answer_consensus}
\end{equation}
TTRV additionally subtracts an entropy penalty computed from the answer distribution. Both terms depend only on the final answers, so responses that share an answer receive the same reward regardless of their reasoning. ViDeR only keeps $p(a_i)$ and modulates it with a visual-dependence score $S_i$.

\section{Methods}

ViDeR consists of two components (\cref{fig:framework}). Attention-guided
visual intervention locates the image regions that a response's reasoning
attends to, occludes them, and re-scores the same response to measure how much
its likelihood drops. The Visual-Dependence Reward aggregates these token-level
drops into a response-level score $S_i$ and uses it to scale the
answer-consensus reward. Responses that share the same final answer therefore
receive different rewards, and responses whose reasoning relies more on the
occluded visual evidence are reinforced.

\subsection{Attention-Guided Visual Intervention}
\label{sec:visual_intervention}
The final answer does not reveal whether the reasoning that produced it used the image or language priors, and intermediate reasoning steps have no direct supervision. We therefore probe this reliance through intervention: if a response relies on particular visual evidence, removing that evidence should lower the likelihood of its own reasoning, whereas a response driven by language priors should be largely unaffected. Attention indicates where the model looks during reasoning, which provides a basis for locating this evidence.

Let $\mathcal{P}$ denote the image-token positions. For each reasoning token $t\in\mathcal{T}_i$, $A_{i,t,v}$ is its attention to image token $v\in\mathcal{P}$, averaged over heads and full-attention layers.
We accumulate attention across the reasoning trace and select the
most-attended visual locations:
\begin{equation}
    u_{i,v}=\sum_{t\in\mathcal{T}_i} A_{i,t,v},
    \qquad
    \mathcal{H}_i=\operatorname{TopK}_{v\in\mathcal{P}}(u_{i,v},k),
    \label{eq:guided_region}
\end{equation}
where $k=\operatorname{round}(\rho|\mathcal{P}|)$ and $\rho$ is the
occlusion ratio. Mapping $\mathcal{H}_i$ to image regions gives an intervened image
$\widetilde I_i=\operatorname{Occ}(I,\mathcal{H}_i)$, with the selected
regions replaced by black pixels and the remaining visual context
preserved. Since $\mathcal{H}_i$ is derived from the
attention of response $o_i$, each response is paired with its own
intervened image, and the occluded region reflects the evidence that this
reasoning trace used rather than the parts of the image that are merely
visually salient.

We do not regenerate under the intervened image. Instead, we teacher-force the generated response $o_i$ under both images and record the per-token log-probabilities. This measures how likely the model would be to produce the same reasoning if the evidence were hidden:
\begin{equation}
    \begin{aligned}
        \ell^{I}_{i,t}
            &=\log\pi_{\theta_{\mathrm{old}}}(o_{i,t}\mid I,q,o_{i,<t}),\\
        \ell^{\mathrm{occ}}_{i,t}
            &=\log\pi_{\theta_{\mathrm{old}}}(o_{i,t}\mid\widetilde I_i,q,o_{i,<t}),
    \end{aligned}
    \label{eq:fixed_response_contrast}
\end{equation}
Both passes share the question $q$ and the prefix $o_{i,<t}$ and differ only in
the image, and neither requires a new rollout. Tokens that depend on the removed evidence become less likely under the occluded image. Therefore, the log-likelihood difference can serve as an indirect measure of the response's reliance on that evidence.
Then, we keep the positive part of the difference:
  \begin{equation}
      d_{i,t}=\max\!\left(\ell^{I}_{i,t}-\ell^{\mathrm{occ}}_{i,t},\,0\right),
      \label{eq:positive_drop}
  \end{equation}
where $d_{i,t}$ is intended to quantify the support that $\mathcal{H}_i$ provides to token $o{i,t}$. If the likelihood of a token increases under occlusion, it has not received support from the occluded region, and we set its value to zero to prevent interference.

\subsection{Visual-Dependence Reward}
\label{sec:visual_reward}
The Visual-Dependence Reward rescores each response according to how much its
reasoning depends on the occluded evidence. It adjusts the answer-consensus
reward so that responses relying more on visual
evidence receive higher scores, even when they share the same final answer.
Concretely, it aggregates the token-level drops $d_{i,t}$ into a
response-level score $S_i$, forming final rewards.

The drops $d_{i,t}$ in \cref{eq:positive_drop} are log-probability ratios, so they measure relative change rather than absolute change: a token that falls from $0.9$ to $0.85$ and one that falls from $0.1$ to $0.05$ both lose $0.05$ in probability, but the second is halved. Moreover, since the probability under occlusion can approach zero, $d_{i,t}$ is unbounded, and a raw average would let a few such tokens dominate the score. We therefore map each drop to $[0,1)$ and scale it by an anchor that measures how close the sampled token is to the model's top choice, which gives the token score:
\begin{equation}
    \begin{aligned}
        \alpha_{i,t}&=\exp\!\left(\ell^{I}_{i,t}-\ell^{\max}_{i,t}\right),\\
        s_{i,t}&=\alpha_{i,t}\,\frac{d_{i,t}}{1+d_{i,t}},
    \end{aligned}
    \label{eq:token_visual_dependence}
\end{equation}
where $\ell^{\max}_{i,t}$ is the largest next-token log-probability under the original image and the same prefix. The anchor $\alpha_{i,t}\in(0,1]$ equals $1$ when the sampled token is the model's top choice and decays as it departs from that choice.

These token scores are then combined into one response-level score.
A token that attends mainly to regions outside $\mathcal{H}_i$ changes little when $\mathcal{H}_i$ is occluded, so its drop is small.
A plain average would let these small values pull $S_i$ down, resulting in scores that would no longer reflect how much the response relies on the removed evidence. We therefore weight each token by its attention mass on $\mathcal{H}_i$, which makes $S_i$ an average over the tokens that pay attention to the removed evidence.
\begin{equation}
    w_{i,t}=\sum_{v\in\mathcal{H}_i} A_{i,t,v},
    \qquad
    S_i=\frac{\sum_{t\in\mathcal{T}_i}w_{i,t}s_{i,t}}
              {\sum_{t\in\mathcal{T}_i}w_{i,t}+\epsilon},
    \label{eq:weighted_visual_dependence}
\end{equation}
where $\epsilon>0$ stabilizes the normalization.

Finally, we scale the answer-consensus reward by the response-level score to obtain the visual-dependence reward $R_i$, which is optimized with GRPO~\cite{GRPO}:
\begin{equation}
    R_i=p(a_i)\bigl(1+S_i\bigr).
    \label{eq:visual_grounded_reward}
\end{equation}
The reward operates at two levels. Across answers, a minority answer whose reasoning relies more on its attended regions can be ranked above a majority answer whose reasoning barely changes without them, so the update is not determined by vote count alone. Among responses that yield the same answer, where answer frequency assigns identical rewards, it favors those whose reasoning relies more on the occluded evidence. A trace whose likelihood barely changes when its attended regions are removed shows little reliance on them, which is the behavior of reasoning driven by language priors and which answer-consensus rewards alone cannot penalize.

\section{Experiments}

\begin{table*}[!t]
\centering
\caption{Comparison across eight visual benchmarks. Bold indicates the best available
result for each backbone and benchmark. The first row in each group reports the pretrained backbone.}
\label{tab:main_results}
\begingroup
\small
\setlength{\tabcolsep}{4pt}
\renewcommand{\arraystretch}{1.0}
\resizebox{\textwidth}{!}{%
\begin{tabular}{l*{9}{c}}
\toprule
\textbf{Method} &
\textbf{MathVerse} &
\textbf{MathVista} &
\textbf{SEED} &
\textbf{MME} &
\textbf{RealWorldQA} &
\textbf{CAPTURe} &
\textbf{AI2D} &
\textbf{CRPE} &
\textbf{AVG} \\
\midrule
\textbf{Qwen3.5-0.8B} & 48.65 & 56.33 & 72.76 & 74.43 & 60.65 & 64.35 & 63.76 & 63.81 & 63.09 \\
EM-FT (NeurIPS'25) & 44.57	& 52.65	& 73.43	& 75.89	& \textbf{64.97}	& 65.36	& 64.09	& 68.00	& 63.62  \\
TTRL (NeurIPS'25) & 42.96 &	55.31 & 72.74 & 74.49 & 61.70 & 64.41 & 63.57 & 67.69 & 62.86  \\
TTRV (CVPR'26) & 48.56 & 58.37 & 73.36 & 75.13 & 60.52 & 67.38 & 63.76 & 69.11 & 64.52 \\
\rowcolor{viderlight} \textbf{ViDeR} & \textbf{51.16} & \textbf{60.00} & \textbf{73.80} & \textbf{76.65} & 63.66 & \textbf{67.80} & \textbf{64.53} & \textbf{69.43} & \textbf{65.88} \\
\midrule

\textbf{Qwen3.5-2B} & \textbf{67.34} & 74.69 & 77.02 & 82.23 & 72.42 & 73.74 & 75.48 & 73.18 & 74.51 \\
EM-FT (NeurIPS'25) & 63.42	& 73.67	& 77.09	& 83.19	& 70.72	& 71.19	& \textbf{76.29}	& 73.48	& 73.63  \\
TTRL (NeurIPS'25) &65.24	&76.53	&76.91	&83.12	&72.03	&\textbf{74.67}	&75.72	&73.20	&74.68 \\
TTRV (CVPR'26) & 66.50 & 76.12 & {77.35} & 82.42 & \textbf{72.81} & 59.38 & 75.04 & \textbf{73.81} & 72.93 \\
\rowcolor{viderlight} \textbf{ViDeR} & 67.27 & \textbf{76.73} & \textbf{77.58} & \textbf{83.44} & \textbf{72.81} & 74.14 & 75.74 & 73.80 & \textbf{75.19} \\
\midrule

\textbf{InternVL3-1B} & 28.59 & 45.92 & 50.22 & 65.04 & 43.66 & 61.21 & 46.86 & 41.49 & 47.87 \\
EM-FT (NeurIPS'25) & 36.02	& 48.98	& 66.32	& 76.14	& 50.33	& 70.01	& 43.90	& 58.91	& 56.33  \\
TTRL (NeurIPS'25)  & 35.11	& 49.18	& 63.39	& 74.68	& 53.99	& 69.50	& 48.41	& 54.04	& 56.04 \\
TTRV (CVPR'26) &36.09	&52.45	&65.73	&75.28	&58.82	&73.85	&56.77	&53.33	&59.04 \\
\rowcolor{viderlight} \textbf{ViDeR} &\textbf{39.10}	&\textbf{55.10}	&\textbf{69.23}	&\textbf{78.87}	&\textbf{60.00}	&\textbf{77.76}	&\textbf{60.13}	&\textbf{60.85}	&\textbf{62.63} \\
\midrule

\textbf{InternVL3-2B} & 46.25 & \textbf{66.12} & 68.72 & 79.44 & 66.14 & 57.93 & 70.23 & 67.53 & 65.30 \\
EM-FT (NeurIPS'25) & 49.19	& 62.65	& 72.92	& \textbf{82.80}	& \textbf{67.06}	& 64.67	& 69.75	& \textbf{71.36}	& 67.55  \\
TTRL (NeurIPS'25) & 48.63	& \textbf{66.12}	& 72.71	& 81.22	& 66.80	& 45.05	& 69.60	& 70.87	& 65.13 \\
TTRV (CVPR'26) & 46.25	& 64.90	& 68.54	& 81.35	& 65.62	& 46.19	& 70.30	& 68.68	& 63.98 \\
\rowcolor{viderlight} \textbf{ViDeR} & \textbf{50.32} & 65.31 & \textbf{73.56} & 82.30 & 65.88 & \textbf{70.17} & \textbf{71.86} & 69.85 & \textbf{68.66} \\
\midrule

\textbf{Ministral3-3B} & 32.03 & 48.37 & 61.94 & 68.91 & 45.10 & 71.41 & 61.24 & 60.14 & 56.14 \\
EM-FT (NeurIPS'25) & 22.28	& 50.20	& 63.36	& 71.07	& 50.46	& 73.23	& 61.69	& 59.09	& 56.42 \\
TTRL (NeurIPS'25) & 24.95 & 54.29 & 62.41 & \textbf{72.84} & 50.98 & 71.89 & 60.80 & 60.41 & 57.32 \\
TTRV (CVPR'26) & 41.14 & 55.51 & 59.29 & 70.88 & 48.50 & \textbf{73.67} & 60.13 & \textbf{62.84} & 59.00 \\
\rowcolor{viderlight} \textbf{ViDeR} & \textbf{43.10} & \textbf{56.73} & \textbf{65.99} & 70.56 & \textbf{53.73} & 71.73 & \textbf{62.87} & 62.00 & \textbf{60.84} \\
\midrule

\textbf{Gemma4-E2B (5.1B)} & \textbf{53.54} & \textbf{51.84} & 67.27 & 77.16 & 51.90 & 63.47 & 55.33 & 67.08 & 60.95 \\
EM-FT (NeurIPS'25) & 52.77	& 48.37	& 67.97	& 78.11	& 50.98	& 61.92	& 57.73	& 67.27	& 60.64 \\
TTRL (NeurIPS'25)  & 52.98	& 50.41	& 67.12	& 78.05	& 52.55	& 63.33	& 59.69	& 66.96	& 61.39 \\
TTRV (CVPR'26) & 51.86 & 50.82 & 68.15 & 78.05 & 52.42 & 64.48 & 59.65 & \textbf{67.62} & 61.63 \\
\rowcolor{viderlight} \textbf{ViDeR} & 52.70 & 50.61 & \textbf{70.11} & \textbf{78.49} & \textbf{52.94} & \textbf{66.55} & \textbf{60.17} & 66.50 & \textbf{62.26} \\
\bottomrule
\end{tabular}%
\label{MainResults}
}
\endgroup
\end{table*}

We extensively evaluate ViDeR on eight multimodal benchmarks covering diverse visual reasoning tasks, using models from multiple families with varying sizes.

\paragraph{Experimental Setup.}
We evaluate on eight benchmarks: MathVerse \cite{MATHVERSE}, MathVista \cite{MathVista}, SEED \cite{Seed}, MME \cite{MME}, RealWorldQA \cite{RealWorldQA}, CAPTURe \cite{CAPTURe}, AI2D \cite{AI2D}, and CRPE \cite{CRPE}, using six backbones from four families: Qwen3.5 (0.8B and 2B)~\cite{Qwen35}, InternVL3 (1B and 2B)~\cite{InternVL3}, Ministral 3 (3B)~\cite{Ministral3}, and Gemma 4 E2B~\cite{Gemma4}. We compare against the pretrained backbone, entropy minimization (EM-FT~\cite{EM-FT}), and consensus-based test-time reinforcement learning (TTRL~\cite{TTRL} and TTRV~\cite{TTRV}). Following TTRV~\cite{TTRV}, all methods adapt a separate model for each backbone and benchmark on 20 unlabeled test examples for 2 epochs, sampling 32 responses per prompt and using the first 16 for policy optimization. Each model is evaluated on the full test set, which includes the adaptation examples. Other hyperparameters follow TTRV, and $\rho$ is fixed at 60\% for every backbone and benchmark. All experiments are conducted on $4\times$ NVIDIA A100 GPUs. Further details are provided in the supplementary material.

\subsection{Main Results}
\paragraph{Comparison with entropy-based methods.}
ViDeR outperforms both EM-FT and the pretrained baseline in average score across all six backbones, as shown in \cref{tab:main_results}. Entropy minimization concentrates probability on outputs the model already favors, but these preferences can arise from language priors rather than visual evidence. It can therefore reinforce confident predictions without encouraging better use of the image. In contrast, ViDeR explicitly evaluates visual dependence by measuring how the likelihood of a reasoning trace changes when attended image regions are occluded.
This provides feedback on the image evidence each reasoning trace relies on, giving adaptation a criterion beyond predictive confidence.

\paragraph{Comparison with Consensus-based test-time reinforcement learning methods.}
In \cref{tab:main_results}, the comparison with TTRL and TTRV highlights the value of distinguishing reasoning traces.
ViDeR achieves higher average scores on all six backbones and outperforms TTRV across all eight benchmarks.
These improvements span mathematical reasoning, visual perception, and relational understanding, supporting the broad applicability of response-level visual dependence as an adaptation signal.
Across these tasks, the shared key lies in extracting learning signals from how responses use the image, beyond their consensus on the final answer.
Answer-consensus rewards collapse responses with the same answer into the same reward, even when their reasoning relies on different evidence.
When all sampled responses agree, this reward becomes constant and provides no relative preference among their reasoning traces. In contrast, ViDeR can retain a discriminative signal since these traces differ in visual dependence. This extends reward differentiation from competing answers to the reasoning traces within each answer group.

\subsection{Model Analysis}

\paragraph{Necessity of occlusion.}
A natural question is whether localized occlusion is needed, or whether removing the whole image suffices to measure visual dependence.
TTIQ~\cite{TTIQ}, the closest prior work, follows the latter design: it re-scores responses with the image replaced by an all-zero tensor.
We therefore compare ViDeR with TTIQ under identical adaptation samples, training settings, and evaluation sets.
As shown in \cref{tab:ttiq_comparison}, ViDeR achieves higher average scores on both backbones, with gains of 2.05 points on Qwen3.5-0.8B and 1.74 points on InternVL3-1B.
Removing the whole image measures overall visual dependence and cannot isolate the regions a given reasoning trace attends to.
ViDeR instead occludes only the trace-attended regions while preserving the remaining visual context, so the likelihood change reflects sensitivity to the evidence the response relies on rather than to the presence of the image as a whole.
These results indicate that localized, trace-specific interventions provide a more effective visual-dependence signal than whole-image removal.

\begin{table*}[!t]
    \centering
    \caption{Comparison of ViDeR and TTIQ on eight benchmarks.
    \textsuperscript{\textdagger} marks results from our reimplementation of TTIQ. Bold marks the higher score in each column within each backbone.}
    \label{tab:ttiq_comparison}
    \begingroup
    \small
    \setlength{\tabcolsep}{3.5pt}
    \resizebox{\textwidth}{!}{%
    \begin{tabular}{cl*{9}{c}}
        \toprule
        \textbf{Model} & \textbf{Method} & \textbf{MathVerse} & \textbf{MathVista} & \textbf{SEED} &
        \textbf{MME} & \textbf{RealWorldQA} & \textbf{CAPTURe} &
        \textbf{AI2D} & \textbf{CRPE} & \textbf{AVG} \\
        \midrule
        \textbf{Qwen3.5-0.8B} & TTIQ\textsuperscript{\textdagger} & 48.77 & 58.08 & 72.52 & 74.89 & 61.96 & 64.49 & 63.61 & 66.29 & 63.83 \\
        \rowcolor{viderlight} & ViDeR & \textbf{51.16} & \textbf{60.00} & \textbf{73.80} & \textbf{76.65} &
        \textbf{63.66} & \textbf{67.80} & \textbf{64.53} & \textbf{69.43} & \textbf{65.88} \\
        \midrule
        \textbf{InternVL3-1B} & TTIQ\textsuperscript{\textdagger} & \textbf{40.78} & 46.53 & 67.43 & 76.63 & 58.03 & \textbf{81.98} & 55.72 & 59.98 & 60.89 \\
        \rowcolor{viderlight} & ViDeR & 39.10 & \textbf{55.10} & \textbf{69.23} & \textbf{78.87} &
        \textbf{60.00} & 77.76 & \textbf{60.13} & \textbf{60.85} & \textbf{62.63} \\
        \bottomrule
    \end{tabular}%
    }
    \endgroup
\end{table*}

\begin{figure}[!t]
    \centering
    \captionsetup{skip=3pt}
    \includegraphics[width=\columnwidth]{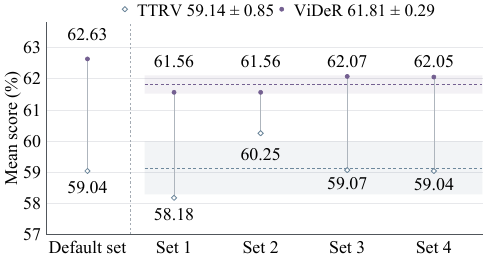}
    \caption{Robustness to adaptation sample selection on InternVL3-1B, evaluated by average scores across eight benchmarks. Dashed lines and shaded bands indicate the mean and $\pm$1 sample standard deviation across the four random subsets (Sets 1--4), excluding the default subset.}
    \label{fig:adaptation_subset}
\end{figure}

\paragraph{Robustness to adaptation sample selection.}
To assess sensitivity to adaptation sample selection, we evaluate InternVL3-1B on the default subset and four random subsets, each with 20 examples per benchmark, using identical samples, training settings, and evaluation sets for TTRV and ViDeR.
As shown in \cref{fig:adaptation_subset}, ViDeR outperforms TTRV on all five subsets; across the four random subsets, it improves the mean score by 2.67\% and reduces the standard deviation to roughly one-third, indicating lower sensitivity to the choice of adaptation samples.

\begin{figure}[!t]
    \centering
    \captionsetup{skip=3pt}
    \includegraphics[width=\columnwidth]{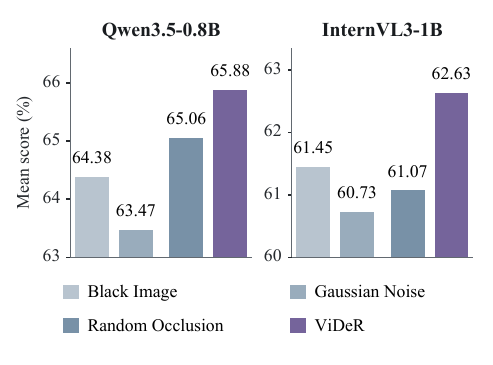}
    \caption{Effect of different visual intervention strategies in ViDeR. Scores represent averages across the eight benchmarks for each backbone.}
    \label{fig:visual_intervention}
\end{figure}

\paragraph{Effect of the visual intervention.}
As shown in \cref{fig:visual_intervention}, attention-guided occlusion achieves the highest mean score on both backbones, although their responses to alternative interventions differ. Notably, the relative performance of whole-image blackening and random occlusion reverses across backbones: random occlusion performs better on Qwen3.5-0.8B, whereas whole-image blackening performs slightly better on InternVL3-1B. This suggests that the effectiveness of coarse-grained visual interventions is backbone-dependent, potentially reflecting differences in how models respond to the removal of visual context. Gaussian noise consistently yields the lowest scores, indicating that generic image corruption may introduce changes unrelated to the visual evidence used in reasoning. More importantly, attention-guided occlusion consistently outperforms random occlusion on both backbones, despite both preserving partial visual context. This highlights that the effectiveness of visual interventions depends not merely on retaining or removing visual information, but on aligning the intervention regions with the evidence attended to by each reasoning trace.

\begin{table}[!t]
    \centering
    \caption{Component ablation of ViDeR. Scores are averages (\%) over eight benchmarks. Ablations replace guided occlusion with random masks of matched coverage and attention weighting with uniform token weights.}
    \label{tab:component_ablation}
    \begingroup
    \small
    \setlength{\tabcolsep}{4pt}
    \renewcommand{\arraystretch}{1.14}
    \begin{tabularx}{\columnwidth}{@{}X*{2}{c}@{}}
        \toprule
        \textbf{Method} & \textbf{Qwen3.5-0.8B} & \textbf{InternVL3-1B} \\
        \midrule
        Pretrained & 63.09 & 47.87 \\
        \midrule
        Only empirical frequency $p$ & 64.25 & 58.97 \\
        Only visual score $S$ & 64.28 & 60.92 \\
        w/o guided occlusion & 65.06 & 61.07 \\
        w/o attention weighting & 64.82 & 61.78 \\
        \rowcolor{viderlight} ViDeR & \textbf{65.88} & \textbf{62.63} \\
        \bottomrule
    \end{tabularx}
    \endgroup
\end{table}

\paragraph{Ablation of attention-guided occlusion.}
\cref{tab:component_ablation} examines the contributions of individual reward components and attention mechanisms in ViDeR. Here, $p$ denotes the reward in \cref{eq:answer_consensus}. The reward ablations reveal complementary roles of answer frequency $p$ and visual dependence $S$. On Qwen3.5-0.8B, both signals achieve nearly identical performance individually, while their combination improves performance by 1.63\% over $p$ and 1.60\% over $S$. On InternVL3-1B, $S$ outperforms $p$ by 1.95\% when used individually, yet combining them yields a further 1.71\% improvement over $S$. These results indicate that the benefits of combining answer-level consensus and trace-level visual dependence persist regardless of their relative individual effectiveness. The attention ablations further reveal two complementary aspects of visual-dependence estimation. Guided occlusion improves performance by 0.82\% and 1.56\% over unguided occlusion on Qwen3.5-0.8B and InternVL3-1B, respectively, highlighting the importance of selecting intervention regions according to each reasoning trace. Attention weighting provides additional gains of 1.06\% and 0.85\% over uniform aggregation, respectively, indicating that the impact of occlusion should be evaluated with respect to the relevance of individual tokens to the intervention. Together, these results demonstrate that both trace-specific region selection and attention-weighted aggregation are important for constructing effective visual-dependence feedback.

\begin{figure}[!t]
    \centering
    \captionsetup{skip=3pt}
    \includegraphics[width=\columnwidth]{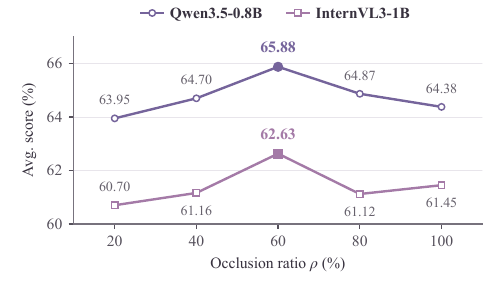}
    \caption{Effect of the occlusion ratio $\rho$ on the average score over
    eight benchmarks. Scores peak at $\rho=60\%$ on both backbones.
    $\rho=100\%$ blacks out the whole image.}
    \label{fig:occ_ratio}
\end{figure}

\paragraph{Effect of the occlusion ratio.}
We vary $\rho\in\{20,40,60,80,100\}\%$ while keeping all other settings fixed, as shown in \cref{fig:occ_ratio}. Both backbones exhibit a consistent non-monotonic trend, achieving their highest average scores at $\rho=60\%$. This shared optimum suggests that an intermediate occlusion ratio provides a favorable balance between removing task-relevant visual evidence and preserving contextual information. With insufficient occlusion, substantial visual evidence may remain accessible, resulting in less informative likelihood changes. Conversely, excessive occlusion may eliminate useful visual context, reducing the specificity of visual-dependence feedback. At $\rho=100\%$, the entire image is replaced with a black image, which prevents the model from accessing any visual evidence. Unlike partial occlusion, this setting cannot isolate the contribution of specific visual regions to individual reasoning traces. In addition, performance at $\rho=40\%$ and $80\%$ remains within approximately 1.5\% of the peak in both backbones, indicating reasonable robustness to moderate variations in the occlusion ratio. These results support $\rho=60\%$ as a practical default setting for all evaluated backbones.

\begin{figure}[!t]
    \centering
    \captionsetup{skip=3pt}
    \includegraphics[width=\columnwidth]{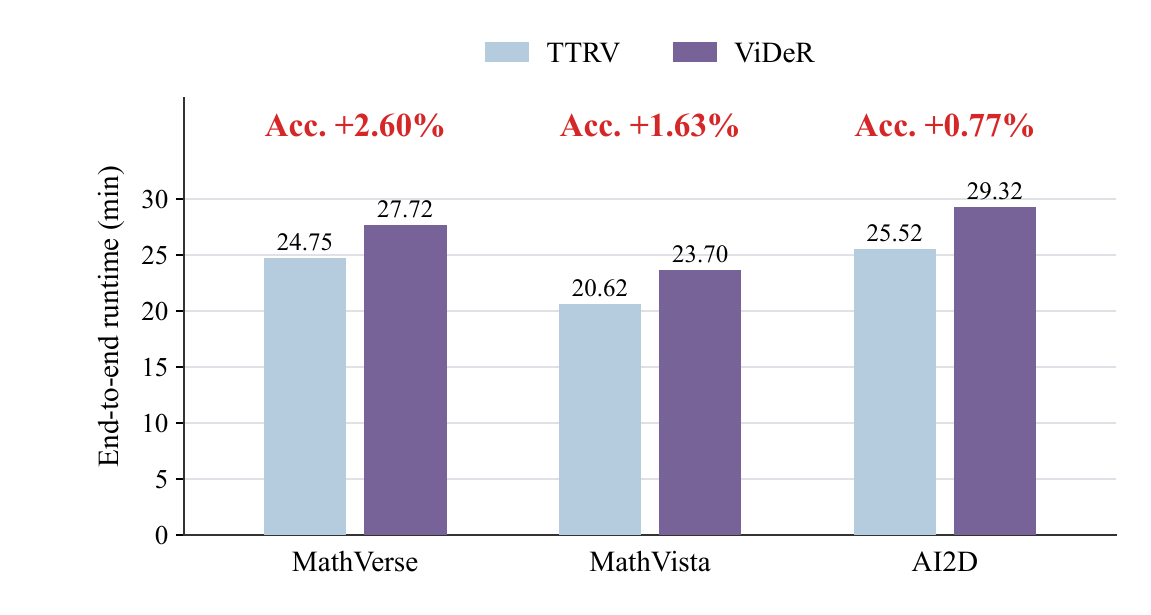}
    \caption{End-to-end runtime and overhead relative to TTRV.}
    \label{fig:efficiency}
\end{figure}

\paragraph{Computational overhead.}
As shown in \cref{fig:efficiency}, ViDeR increases end-to-end runtime relative to TTRV on Qwen3.5-0.8B across the three datasets. This corresponds to approximately three additional minutes and accompanies performance improvements on all three datasets. These results suggest a reasonable performance--runtime trade-off, particularly when predictive performance is prioritized. Inference optimization remains an active research area, and continued advances in engines may further reduce the inference component of end-to-end runtime.

\section{Limitations and failure cases}
ViDeR estimates visual dependence through attention-guided occlusion. This requires an attention-probed forward pass and a re-scoring pass under the occluded image for each response, which increases adaptation cost relative to answer-only reward. In addition, $S_i$ measures reliance on the attended regions, which is distinct from whether that evidence is perceived correctly, and a reward that accounts for both reliance on visual evidence and its correctness is a natural next step. Finally, ViDeR encourages the use of visual information but does not by itself improve question understanding. For example, when asked who is depicted in an Iron Man poster, an adapted model may answer Tony Stark instead of Robert Downey Jr., confusing the character with the actor. Further analyses are provided in the supplementary material.

\section{Conclusion}
We introduced ViDeR to mitigate reliance on language priors in consensus-based test-time reinforcement learning for multimodal reasoning.
Unlike answer-consensus rewards, which treat responses with the same final answer equally, ViDeR distinguishes them by their reliance on visual evidence.
ViDeR measures visual dependence by comparing token probabilities before and after attention-guided occlusion, while keeping each reasoning trace fixed.
This signal is incorporated into the answer-consensus reward to encourage greater reliance on visual evidence during reasoning.
Experiments on eight benchmarks and six backbones show consistent gains over answer-consensus baselines.
Future work will investigate how to improve models' understanding of questions alongside their use of visual evidence.

\section*{LLM Usage Statement}
We used Codex and Claude Code to assist with code development,
improve manuscript wording and clarity, and check LaTeX formatting.
Generative AI was not used to write the initial draft of the
manuscript, create the figures or tables (including the framework
diagram), or generate synthetic data.
All AI-provided questions and suggestions were assessed by the
authors, who wrote and revised the manuscript and take
responsibility for the final content of this work.

{
    \small
    \bibliographystyle{ieeenat_fullname}
    \bibliography{main}
}
\onecolumn
\raggedbottom
\widowpenalty=10000
\vbox{
    \centering
    \Large
    \textbf{\thetitle}\par
    \vspace{0.5em}
    Supplementary Material\par
}
\vspace{1em}

\appendix

This supplementary material provides additional implementation details, analyses, and qualitative results for ViDeR. Specifically, we describe the experimental setup, hyperparameters, and complete adaptation algorithm in \cref{sec:benchmarks}, summarize the six evaluated vision-language backbones in \cref{sec:supp_backbones}, and list the exact prompt templates in \cref{sec:supp_prompts}. We then report extended ablation studies on adaptation set size and visual intervention strategies in \cref{sec:supp_more_results}, followed by qualitative comparisons across all eight benchmarks in \cref{sec:supp_qualitative} and representative failure cases with future directions in \cref{sec:supp_failure}.

\section{Experimental Setup}
\label{sec:benchmarks}

\paragraph{Benchmarks.}
We evaluate on eight benchmarks spanning mathematical reasoning, general multimodal perception, and spatial-relational understanding:
\begin{itemize}
    \item \textbf{MathVista and MathVerse} test mathematical reasoning in visual contexts.
    \item \textbf{AI2D} focuses on scientific diagram understanding and question answering.
    \item \textbf{MME and SEED} assess broader multimodal capabilities, including object recognition, scene understanding, and visual reasoning.
    \item \textbf{RealWorldQA} emphasizes spatial understanding in real-world scenes.
    \item \textbf{CRPE} probes object recognition and relation comprehension.
    \item \textbf{CAPTURe} evaluates spatial reasoning through counting visible and occluded objects arranged in regular patterns.
\end{itemize}
Across all benchmarks, we follow the data filtering protocol of TTRV, with the resulting test-set sizes reported in~\cref{tab:benchmarks}.

\begin{table}[!htbp]
    \centering
    \caption{Benchmark summary and filtered test-set sizes.}
    \label{tab:benchmarks}
    \small
    \begin{tabular*}{0.8\linewidth}{
        @{\extracolsep{\fill}} llr @{}
    }
        \toprule
        Dataset     & Task                 & \#Test Samples \\
        \midrule
        MathVerse   & Math reasoning       & 1,427 \\
        MathVista   & Math reasoning       & 490   \\
        SEED        & Visual understanding & 3,881 \\
        MME         & Visual understanding & 1,576 \\
        RealWorldQA & Real-world reasoning & 765   \\
        CAPTURe     & Visual counting      & 817   \\
        AI2D        & Diagram reasoning    & 2,704 \\
        CRPE        & Relation reasoning   & 7,576 \\
        \bottomrule
    \end{tabular*}
\end{table}

\paragraph{Optimization and Decoding.}
Table~\ref{tab:vider_hyperparameters} lists the full hyperparameter configuration used for ViDeR. For each backbone and benchmark, TTRV and all compared methods share identical adaptation and evaluation splits, as well as the same rollout and optimization settings where applicable. Specifically, we sample $G=32$ responses per prompt at temperature $1.0$ (with top-$p=1.0$ and top-$k$ disabled) and use the first 16 trajectories for GRPO updates, fixing both data-shuffle and rollout seeds to 0. During evaluation, we generate one response per question via greedy decoding.

\begin{table}[!htbp]
\centering
\small
\caption{ViDeR adaptation hyperparameters used in the main experiments.}
\label{tab:vider_hyperparameters}
\begin{tabularx}{0.8\linewidth}{@{}XX@{}}
\toprule
\textbf{Hyperparameter} & \textbf{Value} \\
\midrule
Unlabeled examples / epochs & 20 / 2 \\
Rollouts per prompt ($G$) & 32 sampled (first 16 used) \\
Rollout / actor batch size & 4 prompts / 16 trajectories \\
Max prompt / response length & 7{,}524 / 1{,}024 tokens \\
Sampling (temp. / top-$p$ / top-$k$) & $1.0$ / $1.0$ / Disabled \\
Occlusion ratio $\rho$ / stabilizer $\epsilon$ & $60\%$ / $10^{-6}$ \\
\midrule
Optimizer & AdamW \\
Learning rate & $5\times 10^{-7}$ (cosine decay) \\
Max gradient norm & $1.0$ \\
GRPO updates per rollout batch & 1 \\
GRPO clip interval (dual-clip) & $[0.8, 1.2]$ ($3.0$) \\
KL loss coeff. / reward KL & $10^{-3}$ / Disabled \\
\bottomrule
\end{tabularx}
\end{table}

\paragraph{Evaluation Metrics.}
Following the adaptation protocol, each adapted model is evaluated on the test set, including the 20 adaptation examples. We report final-answer accuracy on all benchmarks except CAPTURe, where we use the mean $1-\mathrm{sMAPE}$ score. The AVG column reports the unweighted average across all eight benchmarks.

\pagebreak[2]
\begin{samepage}
\paragraph{Algorithm.}
\Cref{alg:vider} summarizes the full ViDeR adaptation procedure. For each prompt, we sample $G=32$ responses from the rollout policy $\pi_{\theta_{\mathrm{old}}}$ and compute the empirical answer frequencies $p(a_i)$ across all $G$ responses to stabilize the consensus estimate. Following TTRV's update budget, we then take the first $M=16$ responses for attention-guided occlusion, visual-dependence scoring, and GRPO optimization. Both scoring passes evaluate the same fixed response under $\pi_{\theta_{\mathrm{old}}}$ without regenerating it, differing only in whether the attended image regions $\mathcal{H}_i$ are occluded. The complete set of optimization hyperparameters is listed in Table~\ref{tab:vider_hyperparameters}.
\par
\end{samepage}

\begin{algorithm}[!htbp]
\caption{ViDeR test-time adaptation}
\label{alg:vider}
\small
\algrenewcommand\algorithmicindent{1em}
\begin{algorithmic}[1]
\Require Pretrained policy $\pi_{\theta_0}$, unlabeled adaptation set $\mathcal{D}$, epochs $E$, rollout size $G$, update size $M$, occlusion ratio $\rho$, stabilizer $\epsilon$
\Ensure Adapted policy $\pi_\theta$
\State $\theta \gets \theta_0$
\For{each rollout batch $\mathcal{B}$ over $E$ epochs}
    \State $\theta_{\mathrm{old}} \gets \theta$
    \ForAll{$(I, q) \in \mathcal{B}$}
        \State Sample $G$ responses $\{o_i\}_{i=1}^G \sim \pi_{\theta_{\mathrm{old}}}(\cdot \mid I, q)$
        \State Extract answers $\{a_i\}_{i=1}^G$ and reasoning spans $\{\mathcal{T}_i\}_{i=1}^M$
        \State $p(a_i) \gets \frac{1}{G}\sum_{j=1}^G \mathbf{1}[a_j = a_i]$ for $i = 1, \dots, M$
        \State Let $\mathcal{P}$ be the image-token positions; $k \gets \operatorname{round}(\rho|\mathcal{P}|)$
        \For{$i = 1, \dots, M$}
            \State Forward $o_i$ on $(I, q)$ with $\pi_{\theta_{\mathrm{old}}}$ to get $A_{i,t,v}$, $\ell^I_{i,t}$, $\ell^{\max}_{i,t}$
            \State $u_{i,v} \gets \sum_{t\in\mathcal{T}_i} A_{i,t,v}$, \quad $\mathcal{H}_i \gets \operatorname{TopK}_{v\in\mathcal{P}}(u_{i,v}, k)$
            \State $\widetilde{I}_i \gets \operatorname{Occ}(I, \mathcal{H}_i)$
            \State Forward $o_i$ on $(\widetilde{I}_i, q)$ with $\pi_{\theta_{\mathrm{old}}}$ to get $\ell^{\mathrm{occ}}_{i,t}$
            \ForAll{$t \in \mathcal{T}_i$}
                \State $d_{i,t} \gets \max(\ell^I_{i,t} - \ell^{\mathrm{occ}}_{i,t}, 0)$
                \State $s_{i,t} \gets \exp(\ell^I_{i,t} - \ell^{\max}_{i,t})\,\frac{d_{i,t}}{1 + d_{i,t}}$
                \State $w_{i,t} \gets \sum_{v\in\mathcal{H}_i} A_{i,t,v}$
            \EndFor
            \State $S_i \gets \frac{\sum_{t\in\mathcal{T}_i} w_{i,t} s_{i,t}}{\sum_{t\in\mathcal{T}_i} w_{i,t} + \epsilon}$
            \State $R_i \gets p(a_i)(1 + S_i)$
        \EndFor
    \EndFor
    \State Update $\theta$ with GRPO using $\{(o_i, R_i)\}_{i=1}^M$ per prompt
\EndFor
\State \Return $\pi_\theta$
\end{algorithmic}
\end{algorithm}

\section{Backbones}
\label{sec:supp_backbones}
We select six pretrained models from four open-weight vision-language model families widely used in multimodal research and local deployment. These models differ in how they encode images and how their decoders attend to visual tokens, allowing us to test ViDeR across architectures that process visual evidence differently.

\paragraph{Qwen3.5~\cite{Qwen35}.}
Alibaba's Qwen family is a widely used base for community-developed models. We use the 0.8B and 2B Qwen3.5 checkpoints, which support native image input and share a 24-layer decoder layout: in each group of four layers, three use Gated DeltaNet and one uses gated full self-attention. This pair allows us to compare model sizes under a shared attention architecture.

\paragraph{InternVL3~\cite{InternVL3}.}
OpenGVLab's InternVL is an established family for visual understanding and was used in the original TTRV study. Both InternVL3-1B and InternVL3-2B use the InternViT-300M vision encoder with $448\times448$ image tiles, paired with a two-layer MLP that projects visual features into Qwen2.5 decoders initialized from the 0.5B and 1.5B variants, respectively. This pair allows us to compare decoder sizes with a shared vision encoder.

\paragraph{Ministral~3~\cite{Ministral3}.}
Ministral~3 is Mistral AI's compact open-weight multimodal series for local and edge deployment. We evaluate Ministral~3 3B Instruct, the smallest instruction-tuned checkpoint in the series, which pairs a 410M-parameter Pixtral-style vision encoder with a dense decoder using grouped-query attention.

\paragraph{Gemma~4~\cite{Gemma4}.}
Google DeepMind's Gemma family has a large ecosystem of community variants. We use the instruction-tuned Gemma~4 E2B, which targets on-device multimodal deployment. It pairs a 150M-parameter vision encoder with a decoder that uses four local sliding-window layers per global attention layer. Through per-layer embeddings, E2B has 2.3B effective parameters out of 5.1B total.

\section{Prompt Templates}
\label{sec:supp_prompts}
We use a fixed prompt format across all backbones and benchmarks without dataset-specific prompt tuning. The only difference is whether answer choices are provided: the multiple-choice template is used for the seven multiple-choice benchmarks, and the open-ended variant (omitting the choices) is used for counting on CAPTURe.

\begin{promptbox}{Multiple-Choice Benchmarks}
\phtag{<image>}\\[4pt]
Briefly reason using the visual evidence and the choices. Keep the reasoning concise. Use no more than four short sentences. Put your final answer within \texttt{\textbackslash boxed\{\}}.\\[5pt]
\textbf{Question:} \phtag{\{question\}}\\
\textbf{Choices:}\\
\phtag{\{choices\}}
\end{promptbox}

\begin{promptbox}{CAPTURe (Counting)}
\phtag{<image>}\\[4pt]
Briefly reason using the visual evidence. Keep the reasoning concise. Use no more than four short sentences. Put your final answer within \texttt{\textbackslash boxed\{\}}.\\[5pt]
\textbf{Question:} \phtag{\{question\}}
\end{promptbox}

\section{More Experimental Results}
\label{sec:supp_more_results}
\paragraph{Effect of Adaptation Set Size.}
In the main experiments, we fix $N=20$ to match TTRV~\cite{TTRV} without tuning the adaptation budget. \cref{tab:adaptation_set_size} reports results for $N \in \{1, 10, 20, 50, 100\}$ on InternVL3-1B. Even with $N=1$, ViDeR does not collapse and slightly raises the average score from 47.87\% to 48.76\%. Most of the gain appears by $N=10$ at 61.08\% and $N=20$ at 62.63\%. Increasing the set size to $N=50$ adds only 0.47 points on average at 63.10\% despite using 2.5 times more adaptation examples, so $N=20$ already strikes a reasonable balance between accuracy and cost. At $N=100$, the eight-benchmark average drops to 62.16\%, but this decline is not uniform across tasks. MathVista, SEED, AI2D, and CRPE continue to improve up to $N=100$, while MathVerse, MME, and RealWorldQA peak at $N=20$ and decrease by 1.61 to 2.88 points at $N=100$. Across all seven multiple-choice benchmarks, the average score rises slightly from 60.63\% at $N=50$ to 60.75\% at $N=100$. The drop in the overall average comes almost entirely from CAPTURe, which falls from 80.41\% at $N=50$ to 72.02\% at $N=100$. Because CAPTURe is an open-ended counting task rather than a bounded multiple-choice benchmark, adapting on more unlabeled examples may drift toward wrong consensus counts.

\begin{table}[!htbp]
    \centering
    \small
    \setlength{\tabcolsep}{8pt}
    \caption{Effect of adaptation set size $N$ on InternVL3-1B. Shading marks the main-paper setting ($N=20$), and \textbf{bold} indicates the best score in each column.}
    \label{tab:adaptation_set_size}
    \resizebox{\linewidth}{!}{%
    \begin{tabular}{@{}*{10}{c}@{}}
        \toprule
        $N$ & MathVerse & MathVista & SEED & MME & RealWorldQA
            & CAPTURe & AI2D & CRPE & AVG \\
        \midrule
        Pretrained & 28.59 & 45.92 & 50.22 & 65.04 & 43.66
            & 61.21 & 46.86 & 41.49 & 47.87 \\
        1          & 30.91 & 45.92 & 50.58 & 67.95 & 42.56
            & 62.56 & 46.63 & 42.99 & 48.76 \\
        10         & 38.68 & 53.88 & 66.40 & 78.43 & 57.91
            & 78.97 & 54.92 & 59.44 & 61.08 \\
        \rowcolor{viderlight}
        20         & \textbf{39.10} & 55.10 & 69.23 & \textbf{78.87} & \textbf{60.00}
            & 77.76 & 60.13 & 60.85 & 62.63 \\
        50         & 38.12 & 56.53 & 70.88 & 77.41 & 59.35
            & \textbf{80.41} & 60.24 & 61.85 & \textbf{63.10} \\
        100        & 37.49 & \textbf{57.96} & \textbf{71.32} & 76.71 & 57.12
            & 72.02 & \textbf{61.80} & \textbf{62.82} & 62.16 \\
        \bottomrule
    \end{tabular}%
    }
\end{table}

\paragraph{Effect of Visual Interventions.}
Table~\ref{tab:intervention_detail} reports the per-benchmark results of different visual interventions. ViDeR ranks first in 13 of the 16 evaluations, winning 7 of 8 benchmarks on Qwen3.5-0.8B and 6 of 8 on InternVL3-1B. Across both backbones, Gaussian noise yields the lowest average score and exhibits sharp drops on several benchmarks, such as MME on Qwen3.5-0.8B at 69.04\%, which is 6.78 points below whole-image blackening and 7.10 points below random occlusion. On InternVL3-1B, it also records the lowest scores on MathVista at 50.96\% and CAPTURe at 74.34\%. Because Gaussian noise perturbs every pixel rather than removing selected regions, it may alter token likelihoods in ways that are unrelated to the visual evidence a trace actually relies on. Whole-image blackening avoids pixel-level noise, but removing the entire image discards all visual context at once. As a result, the likelihood drop only reflects reliance on having an image present rather than on the specific regions used by a reasoning trace. Consistent with this limitation, whole-image blackening lags behind ViDeR most clearly on benchmarks that involve localized visual details, trailing by 3.06 points on MathVista and 2.94 points on CAPTURe for Qwen3.5-0.8B, and by 2.61 and 2.02 points on the same benchmarks for InternVL3-1B. Random occlusion preserves partial visual context, yet its masks are sampled independently of the reasoning trace. A random mask may therefore cover irrelevant regions while leaving key evidence visible, which likely contributes to its inconsistent ranking relative to whole-image blackening. Specifically, random occlusion outperforms black images on every benchmark except MathVerse for Qwen3.5-0.8B, whereas it falls behind on MathVerse, MME, AI2D, and CRPE for InternVL3-1B. Selecting occluded regions from each trace's attention keeps the intervention aligned with the regions the reasoning actually attends to, which is consistent with the more stable gains observed across both backbones.

\begin{table}[!htbp]
\centering
\small
\setlength{\tabcolsep}{4.5pt}
\caption{Per-benchmark comparison of different visual intervention strategies on Qwen3.5-0.8B and InternVL3-1B, expanding the average results in Fig.~4 of the main paper. \textbf{Bold} marks the highest score in each column within each backbone.}
\label{tab:intervention_detail}
\begin{tabular}{ll*{9}{c}}
\toprule
Model & Intervention & MathVerse & MathVista & SEED & MME & RealWorldQA & CAPTURe & AI2D & CRPE & AVG \\
\midrule
 & Black Image & 49.68 & 56.94 & 73.51 & 75.82 & 62.22 & 64.86 & 63.76 & 68.23 & 64.38 \\
 & Gaussian Noise & 48.28 & 57.55 & 73.49 & 69.04 & 62.75 & 65.17 & 63.91 & 67.54 & 63.47 \\
 & Random Occlusion & 48.91 & 59.18 & 73.64 & 76.14 & 62.35 & 66.42 & \textbf{65.09} & 68.76 & 65.06 \\
\rowcolor{viderlight}
\cellcolor{white}\multirow{-4}{*}{Qwen3.5-0.8B} & \textbf{ViDeR} & \textbf{51.16} & \textbf{60.00} & \textbf{73.80} & \textbf{76.65} & \textbf{63.66} & \textbf{67.80} & 64.53 & \textbf{69.43} & \textbf{65.88} \\
\midrule
 & Black Image & 37.89 & 52.49 & 67.88 & 76.89 & 60.13 & 75.74 & 58.84 & 61.76 & 61.45 \\
 & Gaussian Noise & 37.59 & 50.96 & 66.73 & 75.79 & 58.76 & 74.34 & 59.62 & \textbf{62.08} & 60.73 \\
 & Random Occlusion & 36.58 & 53.18 & 68.50 & 75.51 & \textbf{60.26} & 76.01 & 57.77 & 60.74 & 61.07 \\
\rowcolor{viderlight}
\cellcolor{white}\multirow{-4}{*}{InternVL3-1B} & \textbf{ViDeR} & \textbf{39.10} & \textbf{55.10} & \textbf{69.23} & \textbf{78.87} & 60.00 & \textbf{77.76} & \textbf{60.13} & 60.85 & \textbf{62.63} \\
\bottomrule
\end{tabular}
\end{table}

\clearpage
\section{Qualitative Examples}
\label{sec:supp_qualitative}
\begin{figure}[H]
    \centering
    \includegraphics[width=\dimexpr0.95\linewidth*1534/2169\relax]{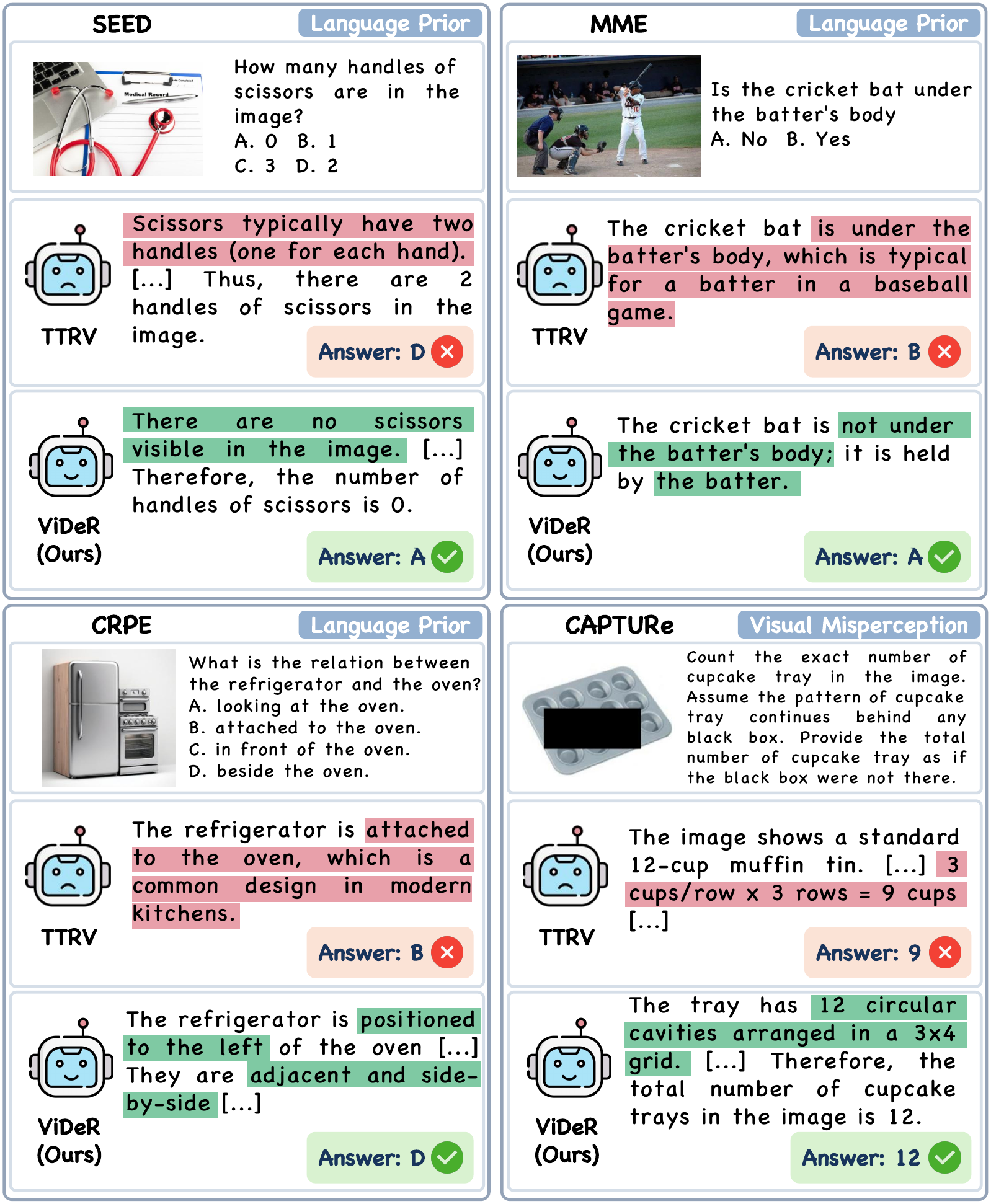}
    \caption{\textbf{Qualitative comparisons on SEED, MME, CRPE, and CAPTURe where language priors and hallucinations lead TTRV to wrong answers.} Red and green highlights mark ungrounded priors or hallucinations and visually grounded reasoning, respectively.}
    \label{fig:qual_wrong}
\end{figure}

\begin{figure}[H]
    \centering
    \includegraphics[width=0.95\linewidth]{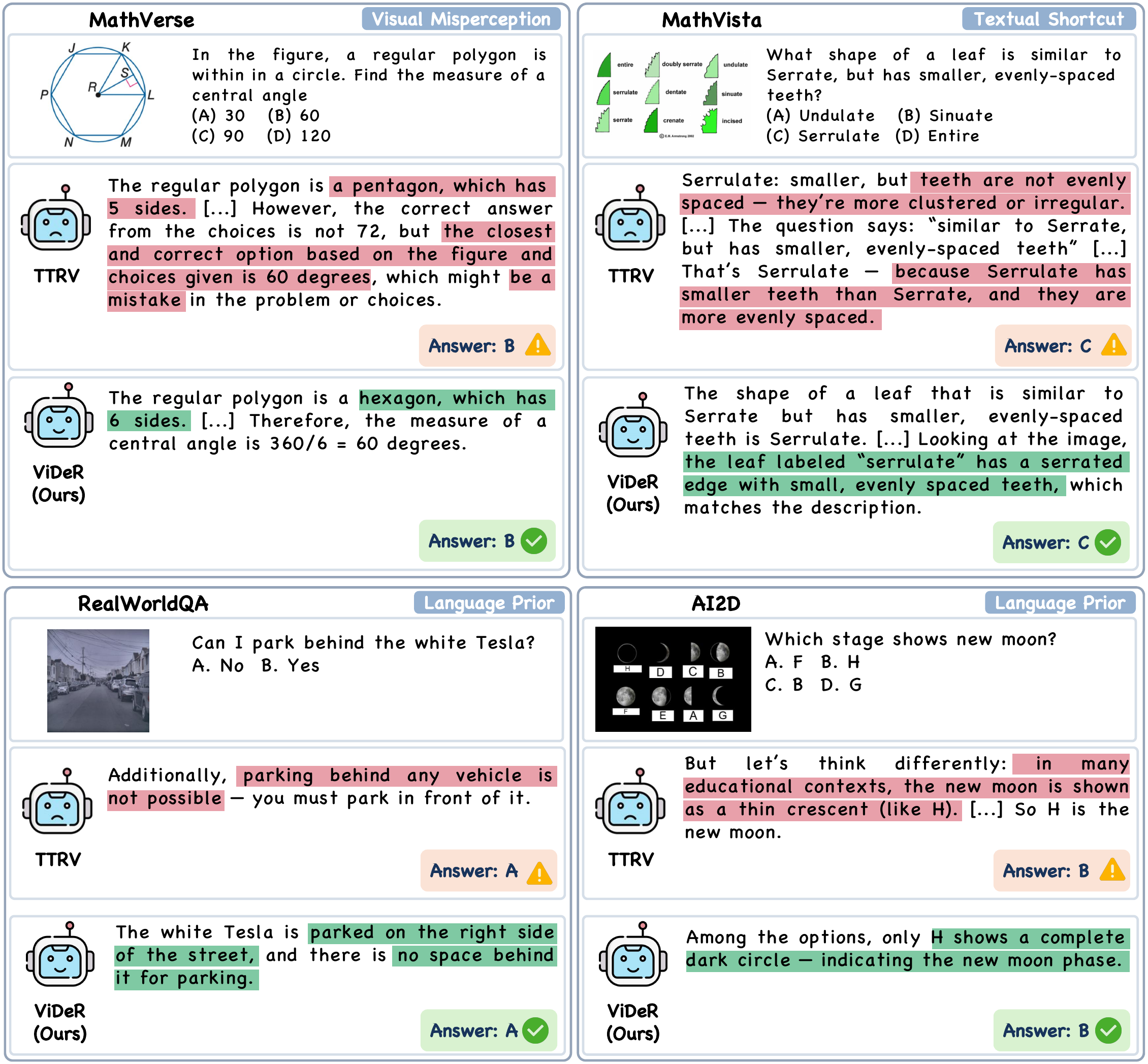}
    \caption{\textbf{Qualitative comparisons on MathVerse, MathVista, RealWorldQA, and AI2D where TTRV gets the right answer for the wrong reasons.} Red and green highlights mark ungrounded priors or hallucinations and visually grounded reasoning, respectively, while orange warning badges indicate correct answers reached via ungrounded shortcuts.}
    \label{fig:qual_ungrounded}
\end{figure}

We present qualitative comparisons across all eight benchmarks to show reasoning failures in TTRV and how ViDeR resolves them. In these examples, TTRV fails through common issues such as language priors, visual misperception, and textual shortcuts. \textbf{Language priors} occur when the model relies on commonsense assumptions or memorized rules instead of checking the image. \textbf{Visual misperception} arises when it misreads visual details such as object counts or geometric shapes. \textbf{Textual shortcuts} happen when it exploits question phrasing or picks the closest option without visual grounding. We group the examples by their final-answer outcomes. In Fig.~\ref{fig:qual_wrong}, these failures lead TTRV to wrong answers, whereas ViDeR inspects the image and derives the right answer from the visible evidence. In Fig.~\ref{fig:qual_ungrounded}, TTRV still selects the right option through ungrounded priors, misperception, or shortcuts, while ViDeR reaches the same answer by grounding its reasoning in the relevant visual details.

\paragraph{Wrong Answers from Ungrounded Reasoning.}
In Fig.~\ref{fig:qual_wrong}, TTRV reaches wrong answers by relying on language priors or misreading the image instead of checking the visual evidence. On SEED, it assumes that scissors have two handles even though no scissors appear in the photo. On MME, it appeals to what is typical in a baseball game to claim that the bat is under the batter's body. On CRPE, it concludes that the refrigerator is attached to the oven because that is a ``common design in modern kitchens.'' On CAPTURe, it first calls the object a 12-cup muffin tin, yet misreads the layout under occlusion and computes $3 \times 3 = 9$. In contrast, ViDeR checks the visible content directly in all four cases. It notes that no scissors are present on SEED, observes that the batter is holding the bat on MME, identifies the side-by-side layout on CRPE, and counts the $3 \times 4$ grid on CAPTURe. These four cases point to a shared weakness of answer-consensus rewards. When a question triggers a strong prior or a familiar scene assumption, multiple sampled rollouts can easily converge on the same plausible claim, so frequency-based rewards end up reinforcing image-insensitive reasoning. Under ViDeR, the likelihood of generic prior statements such as ``scissors typically have two handles'' or ``common design in modern kitchens'' tends to change little when attended image regions are occluded, whereas claims grounded in specific visual details are more sensitive to removing those regions. Weighting answer frequency by visual dependence therefore shifts adaptation away from repeating prior assumptions and toward verifying what is actually visible.

\paragraph{Right Answers from Ungrounded Reasoning.}
In Fig.~\ref{fig:qual_ungrounded}, TTRV selects the right option through ungrounded reasoning. On MathVerse, it mistakes the inscribed hexagon for a pentagon with five sides, sees that $72^\circ$ is absent from the choices, and picks the closest value of $60^\circ$ while assuming the problem contains an error. On MathVista, it arrives at the right label through self-contradictory text reasoning without checking the leaf diagram. On RealWorldQA, it invents a general rule that parking behind any vehicle is impossible rather than inspecting the street scene. On AI2D, it appeals to a supposed convention in educational contexts and misdescribes the dark circle H as a thin crescent. In contrast, ViDeR reaches the same answers by grounding its reasoning in the image. It counts six sides to obtain $360/6 = 60^\circ$ on MathVerse, verifies the leaf labeled ``serrulate'' on MathVista, checks the lack of space behind the white Tesla on RealWorldQA, and identifies H as a complete dark circle on AI2D. These cases show why answer-consensus rewards remain insufficient even when the final answer is right. Because TTRV scores responses solely by answer agreement, shortcut traces based on option proximity, contradictory text, or fabricated rules receive the same reward as visually grounded traces whenever they share the consensus answer. Reinforcing such traces during adaptation encourages reasoning habits that can easily fail on harder inputs. Under ViDeR, responses with the same final answer are separated by their visual-dependence scores. The likelihood of ungrounded claims such as ``parking behind any vehicle is not possible'' or ``in many educational contexts'' tends to change little when attended image regions are occluded, whereas traces that cite specific visual details are more sensitive to occlusion and therefore receive higher rewards within the same answer group.

\section{Failure Cases and Future Directions}
\label{sec:supp_failure}

\begin{figure}[H]
    \centering
    \includegraphics[width=0.92\linewidth]{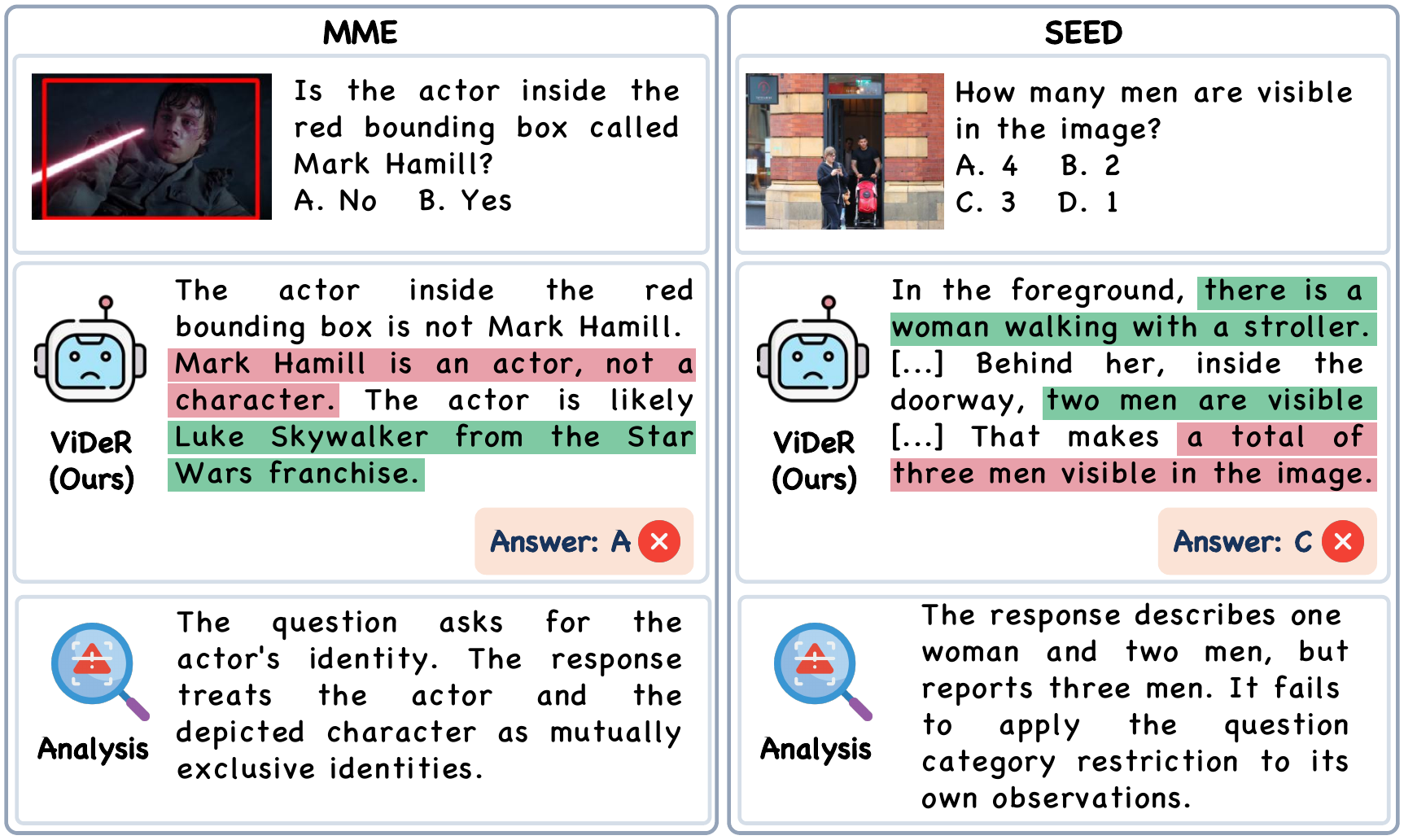}
    \caption{\textbf{Representative failure cases of ViDeR on MME and SEED.} Green highlights indicate accurate visual observations and red highlights mark reasoning errors caused by misinterpreting the question or ignoring its category constraint.}
    \label{fig:qual_failure}
\end{figure}

\paragraph{Failure Cases.}
Following Section~6 of the main paper, Fig.~\ref{fig:qual_failure} illustrates two representative failure modes of ViDeR where the model grounds its description in the image yet still produces a wrong answer. The first mode is misinterpreting the intent of the question. Here, the model identifies the visual subject accurately, but misunderstands what role or attribute the question asks about. In the MME example, ViDeR recognizes Luke Skywalker from the Star Wars franchise, yet denies that the actor is Mark Hamill by treating the actor and the character as mutually exclusive identities. The second mode is failing to enforce question constraints when aggregating visual observations. In this case, the model extracts accurate visual details, but drops a key restriction from the question when forming its final answer. In the SEED example, ViDeR correctly describes one woman with a stroller in the foreground and two men in the doorway, yet counts all three people as men in the conclusion. Both failure modes reflect a boundary of visual-dependence rewards. Since ViDeR estimates visual dependence by occluding attended image regions, a reasoning trace that cites genuine visual details can still experience a likelihood drop under occlusion and receive a high visual-dependence score, even when it misapplies those details to the question. Image occlusion verifies whether a trace relies on the visual input, but it does not check whether the trace respects the semantics of the question or applies its constraints consistently.

\paragraph{Future Directions.}
These observations point to two directions for future work. One direction is to extend counterfactual intervention from the image to the question itself, for instance by measuring how a reasoning trace responds when key constraint words such as ``actor'' or ``men'' are altered. Another direction is to explore label-free consistency signals that verify whether a model's final answer agrees with its own intermediate visual observations under the question's constraints. Jointly accounting for visual dependence, question intent, and logical consistency will be an important step toward more reliable test-time reinforcement learning for VLMs.

\end{document}